\documentclass{article}

\usepackage{PRIMEarxiv}
\usepackage{amsmath}
\usepackage[utf8]{inputenc} 
\usepackage[T1]{fontenc}    
\usepackage{hyperref}       
\usepackage{url}            
\usepackage{booktabs}       
\usepackage{amsfonts}       
\usepackage{nicefrac}       
\usepackage{microtype}      
\usepackage{lipsum}
\usepackage{fancyhdr}       
\usepackage{graphicx}       
\usepackage{multirow}
\usepackage{mathtools}
\usepackage{booktabs}  
\usepackage{makecell}  
\usepackage{colortbl}
\usepackage[inkscapelatex=false]{svg}
\graphicspath{{media/}}     

\title{Estimating Pasture Biomass from Top-View Images: \\A Dataset for Precision Agriculture}

\usepackage{authblk}

\author[1]{Qiyu Liao}
\author[1]{Dadong Wang}
\author[2]{Rebecca Haling}
\author[1]{Jiajun Liu}
\author[1]{Xun Li}
\author[3]{Martyna Plomecka}
\author[4]{Andrew Robson}
\author[4]{Matthew Pringle}
\author[5]{Rhys Pirie}
\author[5]{Megan Walker}
\author[5]{Joshua Whelan}

\affil[1]{Data61, CSIRO}
\affil[2]{Agriculture and Food, CSIRO}
\affil[3]{Google}
\affil[4]{University of New England}
\affil[5]{Meat \& Livestock Australia}

\begin{document}
\maketitle

\footnotetext[1]{Corresponding authors: \texttt{\{qiyu.liao\}@csiro.au}}

\footnotetext[2]{Data Availability and Acknowledgement: The Image2Biomass Pasture Innovation Challenge dataset is derived from data originally captured under \href{https://www.mla.com.au/contentassets/cf6981b49e42475bb7b26c5b6089a305/b.gsm.0010_final_report.pdf}{‘B.GSM.0010 - Tools for real time biomass estimation in pastures’}. The authors acknowledge the support of FrontierSI (previously known as the Cooperative Research Centre for Spatial Information) and matching funds provided by the Australian Government for this project. The annotated dataset discussed here is publicly available under a Creative Commons Attribution-ShareAlike 4.0 International (CC BY-SA 4.0) license through the Kaggle competition platform.}

\begin{abstract}

Accurate estimation of pasture biomass is important for decision-making in livestock production systems. Estimates of pasture biomass can be used to manage stocking rates to maximise pasture utilisation, while minimising the risk of overgrazing and promoting overall system health. We present a comprehensive dataset of 1,162 annotated top-view images of pastures collected across 19 locations in Australia. The images were taken across multiple seasons and include a range of temperate pasture species. Each image captures a 70cm × 30cm quadrat and is paired with on-ground measurements including biomass sorted by component (green, dead, and legume fraction), vegetation height, and Normalized Difference Vegetation Index (NDVI) from Active Optical Sensors (AOS). The multidimensional nature of the data, which combines visual, spectral, and structural information, opens up new possibilities for advancing the use of precision grazing management. The dataset is released and hosted in a Kaggle competition that challenges the international Machine Learning community with the task of pasture biomass estimation. The dataset is available on the official Kaggle webpage: \url{https://www.kaggle.com/competitions/csiro-biomass}

\end{abstract}

\section{Introduction}
\label{sec:intro}

Accurate estimation of pasture biomass is essential for profitable and sustainable grazing management. It supports agricultural productivity while preserving ecosystem integrity. Currently, producers are still heavily reliant on subjective visual assessments to optimise feed allocation efficiency. Inefficiencies resulting from suboptimal management practices can result in significant economic losses and contribute to environmental degradation, including soil erosion, biodiversity loss, and carbon depletion \cite{henry2012potential}. Research shows that improved pasture management through accurate biomass estimation can boost farm profitability by up to 10\% in Australian beef and sheep enterprises. Based on reported pricing in 2018, this translates to gross margin increases of approximately \$96 per hectare for sheep and \$52 per hectare for cattle operations \cite{trotter2018biomass}. In addition to economic benefits, optimized grazing pressure also creates a twin benefit of improved environmental outcomes \cite{trotter2018biomass}.

The challenge of biomass measurement is global in scope. Grazing systems cover roughly 25\% of the Earth’s land surface (approximately 3.4 billion hectares) \cite{asner2004grazing} and manage about 50\% of Australia’s landmass (380 million hectares) \cite{abs2024land}. These vast grassland ecosystems play a vital role in carbon sequestration, water cycle regulation, and biodiversity conservation. Traditional biomass measurement methods \cite{edwards2011pasture, rayburn2003falling, trotter2010evaluating, teal2006season, freeman2007plant, schaefer2016combination} are typically destructive, time-consuming, expensive, or require specialized expertise, limiting their widespread adoption in sustainable grazing systems.

Recent advances in computer vision and artificial intelligence offer transformative opportunities to overcome these limitations. By automating the analysis of pasture images, AI models can estimate biomass more precisely and even at the species level. This reduces the need or effort required for manual calibration while providing an enabling data layer for assessment of pasture genotypic composition and phenological stages – allowing for more accurate derivation of quality attributes that support both livestock nutrition and ecosystem health. Put simply, accurate, real-time biomass estimation at a species level could facilitate adaptive grazing strategies that help to maintain optimal pasture conditions, prevent overgrazing, and support natural regeneration cycles.

\begin{figure*}[t!]
	\centering
    \includegraphics[width=\linewidth]{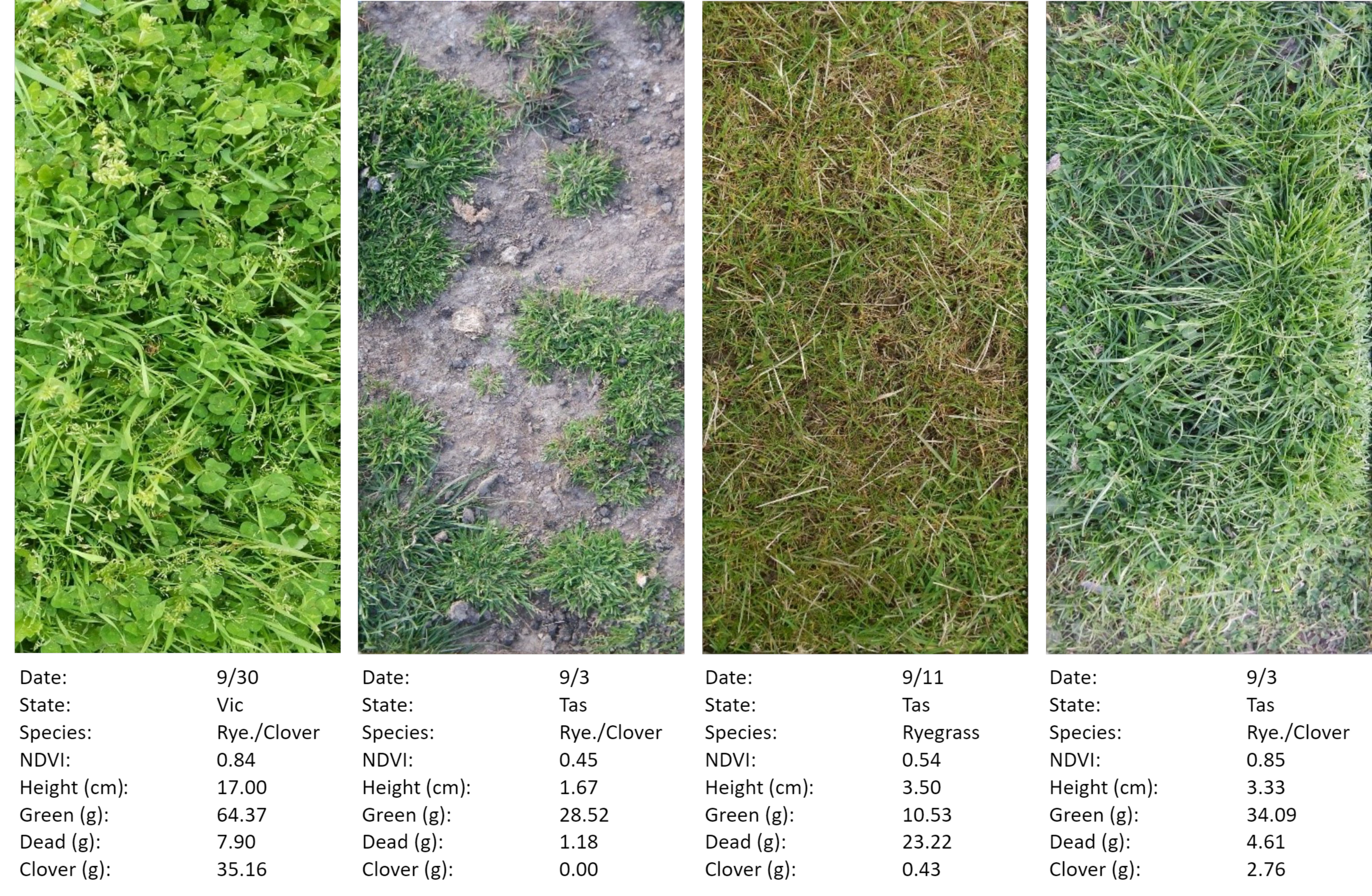}
	\caption{Sample images from the training set and their labels. Images cropped to show the internal section of the 70 x 30cm quadrants, and with dry weight of biomass fractions shown (Green, Dead, Clover).}
	\label{fig:intro}
\end{figure*}

This paper introduces a novel dataset specifically designed to advance AI-driven pasture biomass estimation using top-view imagery (shown in Figure \ref{fig:intro})). This contribution addresses a critical gap in publicly available, professionally annotated datasets that integrate visual, spectral, and structural information for pasture assessment. This enables the development of technologies that support both agricultural productivity and environmental sustainability.

\section{Related Work}
\label{sec:RelatedWork}

\subsection{Traditional Biomass Measurement Methods}

Current approaches to estimating pasture biomass have significant limitations that restrict their practical applications. Destructive methods such as clipping and weighing (cut-dry-weigh method) provide high accuracy through direct measurement but are labor-intensive, time-consuming, and impractical for large-scale or frequent monitoring \cite{edwards2011pasture}. Non-destructive mechanical methods include the rising plate meters, which measure compressed pasture height but require calibration for different pasture types and show variable accuracy with pasture density and moisture content \cite{rayburn2003falling}, and capacitance meters that measure electrical capacitance of vegetation but require complex, pasture-specific calibration with inconsistent results across different conditions \cite{trotter2010evaluating}.

Optical and spectral methods have gained attention because of their potential to provide rapid, nondestructive assessments. Active Optical Sensors (AOS), originally developed for cropping industries to infer nitrogen levels, measure vegetation indices like NDVI by directing light beams onto the canopy and calculating optical reflectance \cite{teal2006season, freeman2007plant, andersson2017estimating}. Research has shown that combining NDVI measurements with height data from LiDAR or mechanical sensors can significantly improve the accuracy of biomass estimation, addressing limitations such as NDVI saturation at high biomass levels, typically when Leaf Area Index (LAI) > 3, and height measurement errors caused by lodging or sparse canopies \cite{schaefer2016combination}. However, the primary challenge remains the development of robust calibration methods suitable for diverse pasture systems \cite{trotter2018biomass}.

Remote sensing approaches using satellite and drone imagery, combined with machine learning, enable large-scale biomass estimation. However, they require technical expertise, significant computational resources, and extensive ground-truth calibration data \cite{edirisinghe2011quantitative}.

\subsection{Computer Vision Approaches and Existing Datasets}

Recent advances in computer vision have shown promising results for agricultural applications, with deep learning models demonstrating success in crop yield prediction and vegetation analysis. However, pasture systems present unique challenges including species diversity, structural complexity, and environmental variability that make them more challenging than monoculture crops.

The most relevant prior work is the GrassClover dataset by Skovsen et al. \cite{228c615fab5c4cf9878ccde8f79b0b4b, skovsen2019grassclover}, which provides 435 images with biomass annotations for grass-clover mixtures. However, the Image2Biomass Pasture Innovation Challenge dataset presented here has several important differentiators compared to the GrassClover dataset. This dataset is larger in scale, offering 1,162 annotated images versus 435, and has additional metadata (species identification, seasonal context, location details, height measurements, and NDVI data). In terms of temporal coverage, the GrassClover dataset is concentrated between May and October, whereas our dataset spans the entire year. Geographically, it is restricted to two locations in Denmark, while our dataset covers 19 sites across four Australian states, representing a wider range of climatic conditions. Additionally, the Image2Biomass Pasture Innovation Challenge dataset includes separate annotations for dead matter. Finally, while the GrassClover images were captured using professional cameras under controlled lighting conditions, our dataset uses diverse consumer-grade cameras in natural lighting environments, enhancing its generalizability and practical applicability for real-world deployment scenarios.

Other approaches have explored synthetic data generation for training augmentation \cite{sapkota2022use}, domain adaptation between ground-level and drone imagery \cite{albert2022unsupervised}, and semi-supervised learning to reduce labeling requirements \cite{albert2021semi}. While these methodological advances are valuable, they remain constrained by the limited size and diversity of the base datasets on which they rely.

Existing agricultural datasets such as CropHarvest (satellite-based crop yield estimation) \cite{tseng2021cropharvest}, PlantNet (species identification) \cite{li2022plantnet}, DeepWeeds (weed detection) \cite{olsen2019deepweeds}, and Agriculture-Vision (aerial field pattern analysis) \cite{chiu2020agriculture} serve distinct purposes within the agricultural domain. However, they lack the ground-level, multi-modal measurements necessary for detailed pasture biomass estimation across diverse environmental conditions and species compositions.

\section{Data Collection}

\subsection{Study Sites and Sampling Protocol}

\begin{figure*}[t!]
	\centering
    \includegraphics[width=\linewidth]{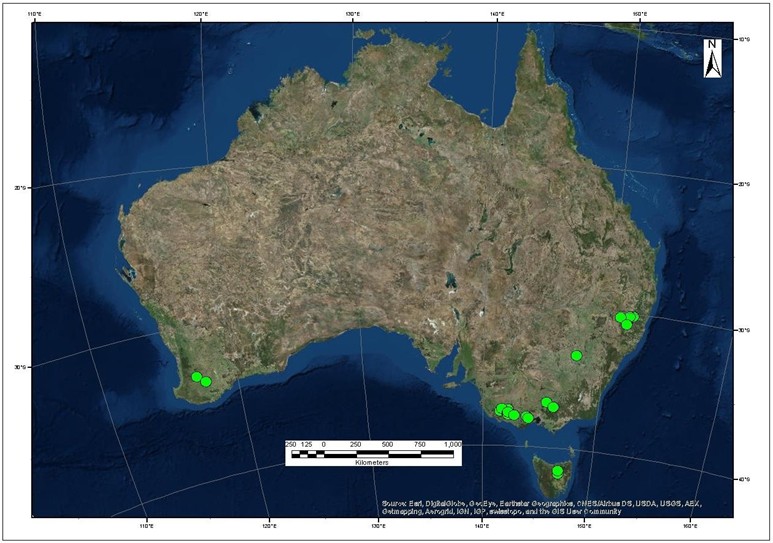}
	\caption{Geographic distribution of sampling sites across Australia \cite{trotter2018biomass}}
	\label{fig:locations}
\end{figure*}

Data collection was conducted across 19 locations spanning four Australian states over a three-year period (2014–2017), representing a wide range of soil, climate, and seasonal conditions across diverse grazing landscapes in the temperate pasture zones of southern Australia.  Multiple sites were selected across New South Wales, Victoria, Tasmania, and Western Australia (illustrated in Figure \ref{fig:locations}).  Sites were strategically selected in collaboration with a network of Meat \& Livestock Australia (MLA)’s Participatory Research Groups to ensure representation of commercially relevant pasture types and management practices.

Each sampling campaign adhered to a rigorous, standardized protocol to ensure data quality and consistency across all sites and seasons. Representative areas within paddocks were carefully selected to avoid edge effects, and transects were established to capture within-paddock variability. Weather conditions were systematically recorded, and sampling was deliberately avoided immediately following rainfall events to prevent soil contamination and maintain consistent measurement conditions.

The data collection sequence began with precise quadrat placement using a standardized 70cm × 30cm metal frame positioned directly on the pasture surface. High-resolution photographs were taken from above using a variety of camera systems (see Table \ref{tab:camera_systems}) to accommodate differing field conditions and research team setups. NDVI measurements were obtained using a GreenSeeker handheld sensor positioned approximately 1m above the center of each quadrat, with an average of 100 readings recorded to account for sensor variability and ensure measurement reliability. Compressed height measurements were collected using a falling plate meter (30cm diameter, 200g weight), with multiple readings taken within each quadrat and averaged to provide representative height values.

Following nondestructive measurements, all vegetation within the quadrat was harvested by cutting to ground level using electric clippers. The harvested material was immediately collected in labeled paper bags to prevent sample loss and maintain traceability. Post-harvest photographs and NDVI readings were taken to document complete vegetation removal and provide quality control data.

\subsection{Laboratory Processing and Quality Control}

All samples underwent standardized laboratory processing within 24 hours of collection to minimize degradation and ensure accurate measurements. Fresh weight was recorded for each complete sample, followed by manual sorting of a minimum 30g subsample into three distinct components: green material (living, non-legume vegetation), dead material (senescent or brown vegetation), and clover, which was separated due to its nutritional significance and nitrogen-fixing properties. This component-wise sorting enables detailed inference of pasture composition and quality, which is essential for accurate livestock nutrition assessment.

All fractions were dried at 70°C for 48 hours following standard protocols, with individual component weights recorded to calculate dry matter content. Final biomass values were calculated per hectare based on the known quadrat area, providing standardized units for model training and evaluation. Rigorous quality control measures included photographic verification of pre- and post-harvest conditions, outlier detection for samples showing evidence of soil contamination or measurement errors, and cross-validation between height and biomass measurements to ensure data integrity.

\subsection{Image Acquisition Systems}

Photography was conducted using a diverse range of camera systems across varying imaging conditions. The equipment used included both consumer-grade and professional-grade devices commonly accessible to researchers and practitioners, thereby enhancing the practical applicability and generalizability of the resulting models.

\begin{table}[h]
\centering
\caption{Camera systems used for image acquisition}
\begin{tabular}{|l|l|}
\hline
\textbf{Manufacturer} & \textbf{Models Used} \\
\hline
Apple & iPhone 4, iPhone 5s \\
\hline
Canon & IXUS 125 HS, COOLPIX AW110 \\
\hline
HTC & 0PJA10 \\
\hline
NIKON & COOLPIX AW110 \\
\hline
OLYMPUS IMAGING CORP. & SP510UZ \\
\hline
Sony & D5833 \\
\hline
\end{tabular}
\label{tab:camera_systems}
\end{table}

Images were captured from above the quadrat at varying heights depending on field conditions and equipment capabilities, with the primary objective of capturing the complete quadrat area at sufficient resolution for detailed vegetation analysis. By employing multiple camera systems and operators, this approach is designed to ensure that the resulting models are not dependent on specific camera characteristics and can generalize effectively across different imaging equipment and field scenarios.

\section{Data Processing and Evaluation}

\subsection{Image Normalization}

Raw images required extensive pre-processing due to variations in perspective and camera positioning across diverse collection sites and equipment types. To establish consistent regions of interest (ROIs), manual annotation of quadrat corners was performed in a counterclockwise order. Special handling procedures were developed for partially visible frames, using edge intersection points to accurately infer quadrat boundaries. Quality control was maintained through systematic visual verification, ensuring accurate boundary identification and consistency across the entire dataset.

An automated pipeline was developed for geometric normalization to ensure consistent image processing regardless of original capture conditions. The pipeline applied perspective correction using affine transformation to convert trapezoidal ROIs into standardized rectangles. Following this, images were resampled to a uniform resolution of 2,000$\times$1,000 pixels, with aspect ratio preservation maintained throughout to prevent geometric distortion that could compromise biomass estimation accuracy. Additionally, automatic orientation alignment was implemented to ensure consistent image orientation across all samples.

\subsection{Data Quality Assurance}

Rigorous quality control measures were implemented to ensure the reliability of the dataset and to exclude samples that could negatively impact model training. Photographic verification involved systematic review of pre- and post-harvest images to confirm complete sampling and identify any procedural errors. Outlier detection protocols flagged samples with unusually high biomass-to-height ratios, which may indicate soil contamination, while NDVI readings outside the typical vegetation range (0.1-0.9) were investigated for potential sensor errors or environmental anomalies.

Cross-validation procedures were used to compare height and biomass measurements, helping to identify inconsistencies, and completeness checks ensured that all required data components were present for each sample. Expert review of species identification was conducted for mixed swards to verify botanical accuracy. Of the initial collection of 3,187 samples, 1,162 passed the comprehensive quality control process. Exclusions were primarily due to incomplete cutting (i.e. vegetation remaining in quadrat after harvest), soil contamination evidenced by unusually high mineral content, measurement errors or missing data components, and non-representative species composition that did not align with target pasture types.

\subsection{Biomass Measurement Standards and Data Integration}

The final biomass measurements represent dry matter content collected within a standardized 70cm $\times$ 30cm quadrat frame, enabling precise component-wise quantification critical for detailed pasture analysis. The measurement protocol produced five key biomass variables:

\begin{itemize}
\item \textit{Dry\_Green\_g}: Green vegetation other than clover (grams)
\item \textit{Dry\_Dead\_g}: Senescent material (grams)
\item \textit{Dry\_Clover\_g}: Clover component (grams)
\item \textit{GDM\_g}: Green dry matter, calculated as the sum of green vegetation and clover (grams)
\item \textit{Dry\_Total\_g}: Total biomass, combining all components (grams)
\end{itemize}

These measurements retain the original quadrat-scale units rather than being extrapolated to per-hectare values, ensuring direct correspondence between image content and biomass labels.

Data integration involved systematically matching image files with corresponding measurement records using unique sample identifiers. The final dataset was formatted as standardized CSV files with consistent column naming conventions to support machine learning model development and ensure reproducibility across research groups and computational environments.

\subsection{Label Structure}

All measurements in the proposed dataset represent dry matter content within the standardized 70cm $\times$ 30cm quadrat frame, ensuring direct correspondence between image content and ground-truth labels. The training set contains complete measurements along with auxiliary metadata to support model development and exploratory data analysis. These auxiliary variables, such as sampling date, location, species composition, NDVI readings, and height measurements are available exclusively in the training set and are excluded from the validation and test sets.

While these auxiliary data can be utilized during model training to potentially improve performance, the evaluation protocol explicitly excludes their use during validation and testing (as shown in Table \ref{tab:dataset_variables}. This design choice reflects real-world deployment scenarios, where such detailed metadata may not be readily available. Consequently, models are expected to rely on visual image features, rather than supplementary measurements that require specialized equipment or extensive field protocols.

\begin{table}[h]
\centering
\caption{Dataset variables and availability across splits}
\begin{tabular}{|l|l|l|l|l|}
\hline
\textbf{Column} & \textbf{Description} & \textbf{Unit} & \textbf{Train} & \textbf{Val/Test} \\
\hline
image\_path & Path to corresponding image & - & \checkmark & \checkmark \\
\hline
Sampling\_Date & Collection date & YYYY-MM-DD & \checkmark & \texttimes \\
\hline
State & Australian state & - & \checkmark & \texttimes \\
\hline
Species & Pasture species (by biomass) & - & \checkmark & \texttimes \\
\hline
Pre\_GSHH\_NDVI & NDVI from GreenSeeker & 0-1 & \checkmark & \texttimes \\
\hline
Height\_Ave\_cm & Average height (falling plate) & cm & \checkmark & \texttimes \\
\hline
Dry\_Green\_g & Green vegetation other than clover & grams & \checkmark & \checkmark \\
\hline
Dry\_Dead\_g & Senescent material & grams & \checkmark & \checkmark \\
\hline
Dry\_Clover\_g & Clover component & grams & \checkmark & \checkmark \\
\hline
GDM\_g & Green dry matter (Green + Clover) & grams & \checkmark & \checkmark \\
\hline
Dry\_Total\_g & Total biomass (all components) & grams & \checkmark & \checkmark \\
\hline
\end{tabular}
\label{tab:dataset_variables}
\end{table}

The five biomass target variables represent the core prediction objectives for machine learning models. \textit{Dry\_Green\_g} quantifies non-legume green vegetation; \textit{Dry\_Dead\_g} measures senescent material crucial for feed quality assessment; \textit{Dry\_Clover\_g} captures the nitrogen-fixing clover component, important for pasture nutrition; \textit{GDM\_g} represents the combined green dry matter, representing the most digestible portion of the pasture; and \textit{Dry\_Total\_g} denotes the total biomass available for livestock consumption. This component-wise breakdown enables nuanced modeling of pasture composition beyond simple total biomass estimation.

\subsection{Evaluation Protocol}

Model performance is evaluated using weighted R$^2$ scores, designed to reflect the relative importance of different biomass components in practical livestock management scenarios. To stabilize variance and improve model evaluation robustness, particularly for variables with high dynamic range or right-skewed distributions, a log-stabilizing transformation is applied to all target variables before computing R$^2$ values:

\begin{equation}
y_{trans} = \log(1 + y)
\end{equation}

where $y$ represents the original biomass measurements in grams. This transformation helps normalize the distribution of biomass values and reduces the impact of extreme outliers on model evaluation.

The final evaluation score is computed as a weighted sum of individual R$^2$ values for each transformed target variable:

\begin{equation}
\text{Final Score} = \sum_{i=1}^{5} w_i \times R^2_i
\end{equation}

where $R^2_i$ is calculated using the log-transformed values, and the weights are distributed as follows: \textit{Dry\_Green\_g} (0.1), \textit{Dry\_Dead\_g} (0.1), \textit{Dry\_Clover\_g} (0.1), \textit{GDM\_g} (0.2), and \textit{Dry\_Total\_g} (0.5). 

This weighting scheme prioritizes total biomass prediction as the primary objective, while still maintaining accuracy across individual components to support detailed analysis of pasture composition. The log transformation ensures that the evaluation metrics are more stable across different scales of biomass measurements and reduces sensitivity to extreme values that may occur in natural pasture environments.

\section{Discussion}

This dataset addresses critical gaps in agricultural computer vision by providing the first large-scale collection of pasture images paired with laboratory-validated, component-wise biomass measurements. Unlike existing datasets that rely on estimated or visually assessed biomass values, our ground-truth measurements undergo rigorous laboratory validation, offering reliable training targets for machine learning models. The detailed breakdown into green, dead, and clover fractions enables nuanced modeling of pasture composition, extending beyond total biomass estimation to support feed quality assessment.

The dataset supports a wide range of research applications, including pasture quality assessment, nitrogen dynamics monitoring in mixed legume-grass systems, and the development of decision support tools for precision grazing management. From a methodological standpoint, it offers opportunities to advance computer vision through multi-task learning, domain adaptation, and uncertainty quantification. 

Reflecting real-world complexity, the dataset presents technical challenges such as extreme biomass density variation, occlusion in dense canopies, and complex spatial patterns in mixed species swards. While comprehensive in scope, limitations include a geographic focus on Australian temperate systems, an emphasis on six major pasture species, and a fixed quadrat size, which may not capture broader landscape-scale patterns.

\section{Conclusion}

We present a comprehensive dataset of 1,162 top-view pasture images paired with laboratory-validated biomass measurements, representing the first large-scale resource to combine visual, spectral, and structural data for precision pasture management. Collected across 19 locations in four Australian states over three years, the dataset captures a wide range of grazing systems and seasonal conditions, with detailed annotations for green vegetation, dead material, and clover fractions.

Released through a Kaggle competition to engage the international machine learning community, this dataset lays the groundwork for developing AI-driven tools aimed at improving livestock production efficiency and supporting sustainable grazing practices across diverse agricultural landscapes.


\bibliography{ref}  

\begin{thebibliography}{10}

\bibitem{henry2012potential}
D.~Henry, J.~Shovelton, C.~de~Fegely, R.~Manning, L.~Beattie, and M.~Trotter.
\newblock Potential for information technologies to improve decision making for the southern livestock industries.
\newblock Technical Report B.GSM.0004, North Sydney, 2012.

\bibitem{trotter2018biomass}
Mark Trotter, Derek Schneider, Andrew Robson, et~al.
\newblock Biomass business ii--tools for real time biomass estimation in pastures.
\newblock 2018.

\bibitem{asner2004grazing}
Gregory~P Asner, Andrew~J Elmore, Lydia~P Olander, Roberta~E Martin, and A~Thomas Harris.
\newblock Grazing systems, ecosystem responses, and global change.
\newblock {\em Annu. Rev. Environ. Resour.}, 29(1):261--299, 2004.

\bibitem{abs2024land}
{Australian Bureau of Statistics}.
\newblock National land account experimental estimates.
\newblock https://www.abs.gov.au/statistics/environment/environmental-management/national-land-account-experimental-estimates/latest-release\#land-use, 2024.
\newblock Accessed: \today.

\bibitem{edwards2011pasture}
C.~Edwards, L.~H. McCormick, and W.~D. Hoffman.
\newblock Pasture skills audit -- a reflection on extension.
\newblock In {\em APEN National Forum: hitting a Moving target Sustaining landscapes, livelihoods and lifestyles in a changing world}, Armidale NSW, 2011. Australasia Pacific Extension Network.

\bibitem{rayburn2003falling}
Ed~Rayburn and John Lozier.
\newblock A falling plate meter for estimating pasture forage mass.
\newblock 2003.

\bibitem{trotter2010evaluating}
M.~G. Trotter, D.~W. Lamb, G.~E. Donald, and D.~A. Schneider.
\newblock Evaluating an active optical sensor for quantifying and mapping green herbage mass and growth in a perennial grass pasture.
\newblock {\em Crop and Pasture Science}, 61:389--398, 2010.

\bibitem{teal2006season}
R.~K. Teal, B.~Tubana, K.~Girma, K.~W. Freeman, D.~B. Arnall, O.~Walsh, and W.~R. Raun.
\newblock In-season prediction of corn grain yield potential using normalized difference vegetation index contribution from the oklahoma agricultural experiment station.
\newblock {\em Agronomy Journal}, 98:1488--1494, 2006.

\bibitem{freeman2007plant}
K.~W. Freeman, K.~Girma, D.~B. Arnall, R.~W. Mullen, K.~L. Martin, R.~K. Teal, and W.~R. Raun.
\newblock By-plant prediction of corn forage biomass and nitrogen uptake at various growth stages using remote sensing and plant height.
\newblock {\em Agronomy Journal}, 99:530--536, 2007.

\bibitem{schaefer2016combination}
Michael~T Schaefer and David~W Lamb.
\newblock A combination of plant ndvi and lidar measurements improve the estimation of pasture biomass in tall fescue (festuca arundinacea var. fletcher).
\newblock {\em Remote Sensing}, 8(2):109, 2016.

\bibitem{andersson2017estimating}
C~Blore.
\newblock Estimating pasture biomass with active optical sensors.
\newblock {\em Advances in Animal Biosciences}, 8(2):754--757, 2017.

\bibitem{edirisinghe2011quantitative}
A~Edirisinghe, MJ~Hill, GE~Donald, and M~Hyder.
\newblock Quantitative mapping of pasture biomass using satellite imagery.
\newblock {\em International Journal of Remote Sensing}, 32(10):2699--2724, 2011.

\bibitem{228c615fab5c4cf9878ccde8f79b0b4b}
S{\o}ren Skovsen, Mads Dyrmann, J{\o}rgen Eriksen, Ren{\'e} Gislum, Henrik Karstoft, and Rasmus \{Nyholm J{\o}rgensen\}.
\newblock Predicting dry matter composition of grass clover leys using data simulation and camera-based segmentation of field canopies into white clover, red clover, grass and weeds.
\newblock In {\em Proceedings of the 14th International Conference on Precision Engineering}. International Society of Precision Agriculture, June 2018.
\newblock International Conference on Precision Agriculture ; Conference date: 24-06-2018 Through 27-06-2018.

\bibitem{skovsen2019grassclover}
Soren Skovsen, Mads Dyrmann, Anders~K Mortensen, Morten~S Laursen, Ren{\'e} Gislum, Jorgen Eriksen, Sadaf Farkhani, Henrik Karstoft, and Rasmus~N Jorgensen.
\newblock The grassclover image dataset for semantic and hierarchical species understanding in agriculture.
\newblock In {\em Proceedings of the IEEE/CVF conference on computer vision and pattern recognition workshops}, pages 0--0, 2019.

\bibitem{sapkota2022use}
Bishwa~B Sapkota, Sorin Popescu, Nithya Rajan, Ramon~G Leon, Chris Reberg-Horton, Steven Mirsky, and Muthukumar~V Bagavathiannan.
\newblock Use of synthetic images for training a deep learning model for weed detection and biomass estimation in cotton.
\newblock {\em Scientific Reports}, 12(1):19580, 2022.

\bibitem{albert2022unsupervised}
Paul Albert, Mohamed Saadeldin, Badri Narayanan, Brian Mac~Namee, Deirdre Hennessy, Noel~E O'Connor, and Kevin McGuinness.
\newblock Unsupervised domain adaptation and super resolution on drone images for autonomous dry herbage biomass estimation.
\newblock In {\em Proceedings of the IEEE/CVF conference on computer vision and pattern recognition}, pages 1636--1646, 2022.

\bibitem{albert2021semi}
Paul Albert, Mohamed Saadeldin, Badri Narayanan, Brian Mac~Namee, Deirdre Hennessy, Aisling O'Connor, Noel O'Connor, and Kevin McGuinness.
\newblock Semi-supervised dry herbage mass estimation using automatic data and synthetic images.
\newblock In {\em Proceedings of the IEEE/CVF International Conference on Computer Vision}, pages 1284--1293, 2021.

\bibitem{tseng2021cropharvest}
Gabriel Tseng, Ivan Zvonkov, Catherine~Lilian Nakalembe, and Hannah Kerner.
\newblock Cropharvest: A global dataset for crop-type classification.
\newblock In {\em Thirty-fifth Conference on Neural Information Processing Systems Datasets and Benchmarks Track (Round 2)}, 2021.

\bibitem{li2022plantnet}
Dawei Li, Guoliang Shi, Jinsheng Li, Yingliang Chen, Songyin Zhang, Shiyu Xiang, and Shichao Jin.
\newblock Plantnet: A dual-function point cloud segmentation network for multiple plant species.
\newblock {\em ISPRS Journal of Photogrammetry and Remote Sensing}, 184:243--263, 2022.

\bibitem{olsen2019deepweeds}
Alex Olsen, Dmitry~A Konovalov, Bronson Philippa, Peter Ridd, Jake~C Wood, Jamie Johns, Wesley Banks, Benjamin Girgenti, Owen Kenny, James Whinney, et~al.
\newblock Deepweeds: A multiclass weed species image dataset for deep learning.
\newblock {\em Scientific reports}, 9(1):2058, 2019.

\bibitem{chiu2020agriculture}
Mang~Tik Chiu, Xingqian Xu, Yunchao Wei, Zilong Huang, Alexander~G Schwing, Robert Brunner, Hrant Khachatrian, Hovnatan Karapetyan, Ivan Dozier, Greg Rose, et~al.
\newblock Agriculture-vision: A large aerial image database for agricultural pattern analysis.
\newblock In {\em Proceedings of the IEEE/CVF Conference on Computer Vision and Pattern Recognition}, pages 2828--2838, 2020.

\end{thebibliography}

\end{document}